\documentclass[final]{cvpr}

\usepackage{times}
\usepackage{epsfig}
\usepackage{graphicx}
\usepackage{amsmath}
\usepackage{amssymb}
\usepackage{paralist}
\usepackage{booktabs}
\usepackage{subcaption}

\usepackage[pagebackref=true,breaklinks=true,colorlinks,bookmarks=false]{hyperref}

\def\cvprPaperID{*****} 
\def\confYear{EarthVision 2021}

\begin{document}

\newcommand{\equationref}[1]{Equation~\ref{#1}}

\newcommand{\reducecaptionspace}{\vspace{-0.3cm}\xspace}
\newcommand{\reducepostfigspace}{\vspace{-0.3cm}\xspace}
\newcommand{\reducespace}{\vspace{-0.3cm}\xspace}

\newcommand{\atl}{\textsc{ATL-SN4}\xspace}
\newcommand{\dfc}{\textsc{DFC19}\xspace}
\newcommand{\sanfernando}{\textsc{ARG}\xspace}
\newcommand{\ner}{\textsc{NER}\xspace}

\newcommand{\rmse}{$RMSE$\xspace}
\newcommand{\rsquared}{$R^2$\xspace}
\newcommand{\tss}{$TSS$\xspace}
\newcommand{\rss}{$RSS$\xspace}

\newcommand{\todo}[1]{\textcolor{red}{TODO: #1}}

\newcommand{\flow}{\textsc{flow}\xspace}
\newcommand{\flowh}{\textsc{flow-h}\xspace}
\newcommand{\flowa}{\textsc{flow-a}\xspace}
\newcommand{\flowha}{\textsc{flow-ha}\xspace}

\newcommand{\specialcell}[2][c]{%
\begin{tabular}[#1]{@{}c@{}}#2\end{tabular}}

\newcommand{\subsubheaderbf}[1]{\mbox{\textbf{#1}\hspace*{2.5mm}}}

\title{Single View Geocentric Pose in the Wild}

\author{Gordon Christie$^{1}$, Kevin Foster$^1$, Shea Hagstrom$^1$, Gregory D. Hager$^2$, Myron Z. Brown$^1$ \\ 
$^1$The Johns Hopkins University Applied Physics Laboratory \\ 
$^2$Department of Computer Science, The Johns Hopkins University \\
{\small \tt \{{gordon.christie},{kevin.foster},{shea.hagstrom},{myron.brown}\}@jhuapl.edu}, \small \tt {hager@cs.jhu.edu} }


\twocolumn[{
\renewcommand\twocolumn[1][]{#1}
\maketitle
\begin{center}
\includegraphics[width=0.95\textwidth]{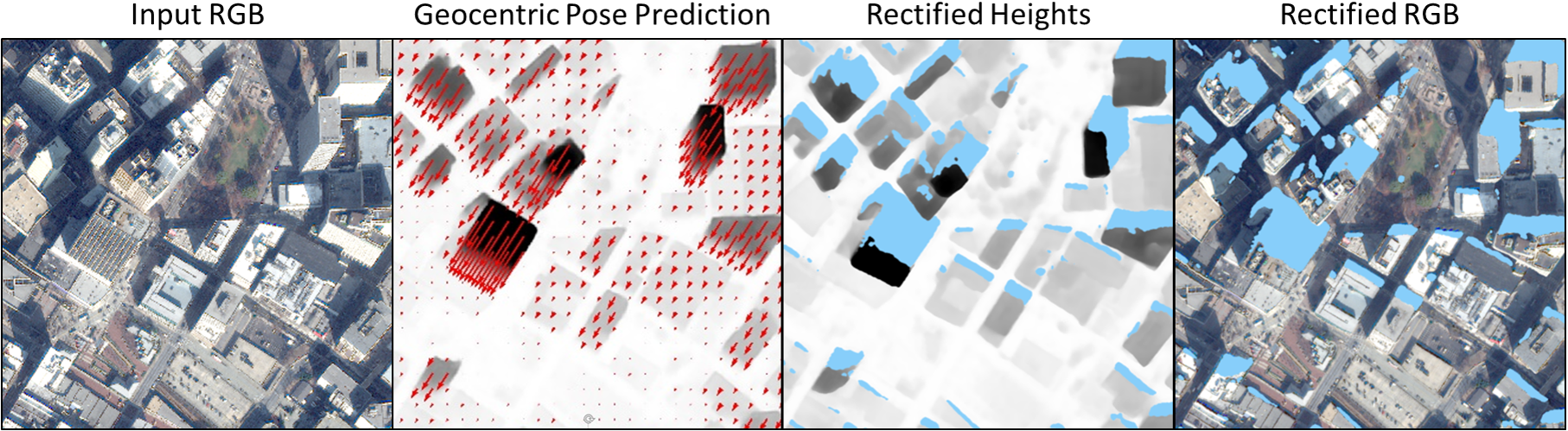}
\captionof{figure}{Our model predicts object heights and vector fields mapping surface features to ground level, enabling feature rectification and occlusion mapping. Darker shades of gray have larger height values, red arrows map surface features to ground level, and occluded pixels are blue.}
\label{fig:geocentric-pose-and-its-uses}
\end{center}

}]

\thispagestyle{empty}
\begin{abstract}
Current methods for Earth observation tasks such as semantic
mapping, map alignment, and change detection rely
on near-nadir images; however, often the first available images
in response to dynamic world events such as natural
disasters are oblique. These tasks are much more difficult
for oblique images due to observed object parallax.
There has been recent success in learning to regress an object's
geocentric pose, defined as height above ground
and orientation with respect to gravity, by training with
airborne lidar registered to satellite images. We present
a model for this novel task that exploits affine invariance
properties to outperform state of the art performance by a
wide margin. We also address practical issues required to
deploy this method in the wild for real-world applications.
Our data and code are publicly available~\footnote{\scriptsize\url{https://github.com/pubgeo/monocular-geocentric-pose}}.
\end{abstract}

\vspace{-0.5cm}

\section{Introduction}

\begin{figure}[h!]
	\centering
	\includegraphics[width=\columnwidth]{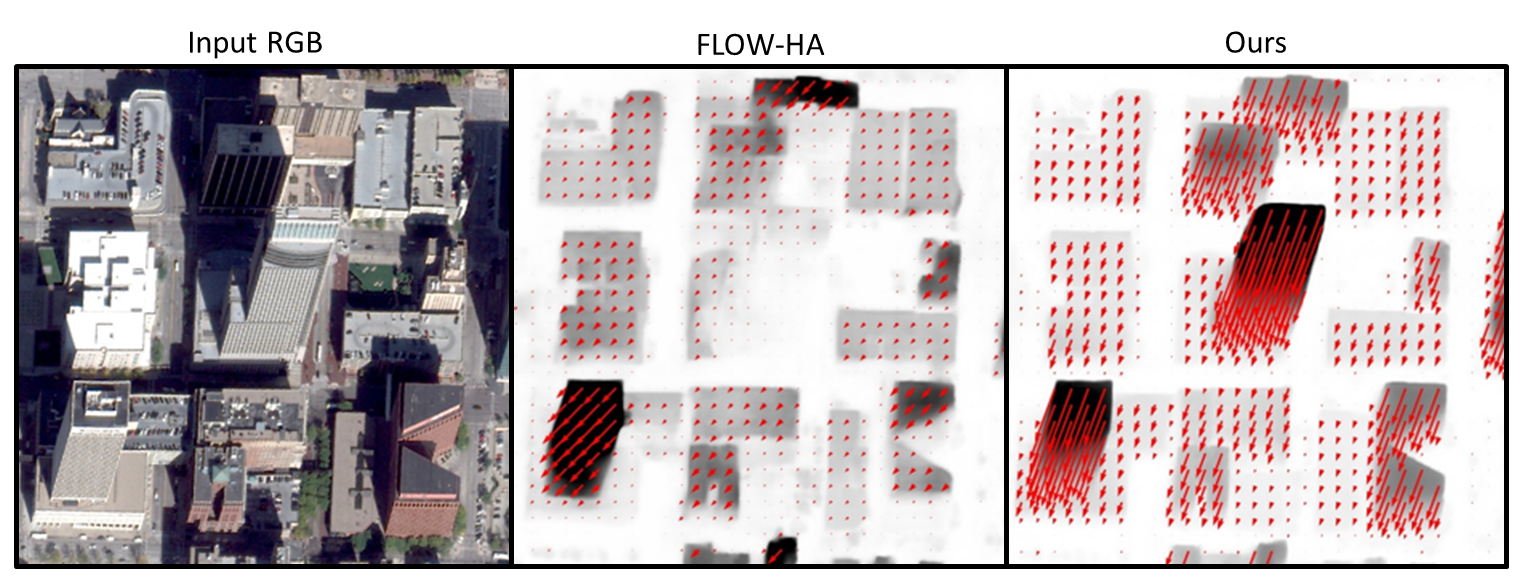}
	\caption{State of the art FLOW-HA from Christie et al. \cite{christie2020learning} under-predicts heights and vector field magnitudes for tall buildings. Our model and augmentations exploit affine invariances to produce accurate predictions. Darker gray is taller, and red arrows map surface features to ground level.
	}
	\label{fig:ours-versus-cvpr20}
\end{figure}

The ability to accurately estimate 3D scene geometry from a single satellite image can dramatically improve automated scene understanding for Earth observation tasks such as urban development monitoring and damage assessment after natural disasters. Most methods for feature mapping \cite{demir2018deepglobe}, map alignment \cite{chen2019autocorrect}, and change detection \cite{doshi2018satellite} require near-nadir images for good performance. Geospatially localizing features in more oblique images is challenging due to above-ground image parallax and occlusion \cite{weir2019spacenet}. These issues can be addressed explicitly with known 3D scene geometry, but such information is generally not known in advance. Until recently, methods for predicting this 3D geometry from a single satellite image also relied on near-nadir imaging geometry for good performance \cite{srivastava2017joint, mou2018im2height, mahmud2020boundary}.
\indent Christie et al. \cite{christie2020learning} recently proposed a method to address this challenge with oblique images by regressing geocentric pose, defined as an object's height above ground and orientation with respect to gravity \cite{gupta2013perceptual}. Their method, supervised by lidar, represents geocentric pose with pixel-level object heights and vector fields mapping surface pixels to ground level. While they demonstrated promising initial results, their model fails to reliably predict heights for tall buildings. In this work, we present a solution that  exploits affine invariances to outperform state of the art by a wide margin (\figref{fig:ours-versus-cvpr20}). Our model produces accurate heights and vector fields even for very tall buildings and produces accurate occlusion maps (\figref{fig:geocentric-pose-and-its-uses}). We also explore practical issues required to deploy our method for real-world applications. Specifically, we make the following contributions: (1) We review affine imaging geometry and exploit invariances to explicitly model the relationship between object heights and the vector field that maps surface features to ground level in an image. (2) To improve prediction of taller building heights, we propose a novel strategy for fast augmentations to synthetically increase the heights of objects by inverting geocentric pose vector fields. (3) We outperform state of the art for height prediction \cite{srivastava2017joint, mou2018im2height, mahmud2020boundary, kunwar2019u, zheng2019pop} and geocentric pose \cite{christie2020learning} and demonstrate accurate predictions even for orthorectified images that violate our affine assumptions. (4) We present the first demonstration of supervising this task without lidar, using only geometry derived from images that can be produced anywhere on Earth. (5) We extend the public dataset from \cite{christie2020learning} to increase geographic diversity and produce consistent train and test sets for public leaderboard evaluation. We make our code available as a strong baseline to promote further research.

\section{Related Work}

\subsection{Monocular Height Prediction}

Deep networks for monocular depth prediction \cite{fu2018deep,zhao2019geometry,godard2019digging,li2018megadepth} have been very successful for applications such as autonomous vehicles where the observed scene is tens of meters from the sensor. Inspired by these successes, recent work has demonstrated similar methods to predict height above ground for Earth observation images where the scene can be hundreds of kilometers from the sensor \cite{mou2018im2height}.

Height and semantic category are complementary intrinsic object attributes. Knowledge of semantic category can constrain height predictions. Trivial examples for remote sensing are ground and water that have no height. Other features such as trees and buildings have known distributions of physically plausible heights. Kunwar \cite{kunwar2019u} and Zheng et al. \cite{zheng2019pop} leveraged semantic cues as priors for height prediction to win the 2019 Data Fusion Contest (DFC19) single-view semantic 3D challenge track \cite{j9229514}. Srivistava et al. \cite{srivastava2017joint} proposed to learn semantics and height jointly with a multi-task deep network. Mahmud et al. \cite{mahmud2020boundary} proposed an especially intriguing model that jointly learns to perform semantic segmentation, height above ground, and a signed distance function from building boundaries. This method in particular relies on near-nadir views where building footprints are consistent with observed roof appearance.

In our work, we explicitly account for oblique imaging geometry, making no assumption of near-nadir views. We do not explicitly reason about semantics, but we still find the height estimates from our geocentric pose model are more accurate than those produced by these state of the art height prediction models that do (\secref{sec:results}). We expect that finer-grained semantic labels for buildings (e.g., commercial vs. residential) or other attributes such as building footprint size may provide more useful complementary information to improve our model. We leave these questions for future work. 

\subsection{Geocentric Pose}

Geocentric pose, defined as height above ground and orientation with respect to gravity, was originally proposed by Gupta el al. \cite{gupta2013perceptual} and used as a feature for object recognition and scene classification. They also proposed a three-channel geocentric pose representation called HHA that encodes horizontal disparity (equivalent to depth), height above ground, and orientation with respect to gravity \cite{gupta2014learning}. This representation has since been used by many works focused on scene understanding \cite{cheng2017locality,qi20173d,lin2017cascaded,park2017rdfnet,wang2018depth,long2015fully,schwarz2018rgb,liu2018see}.

Inspired by the success of the this representation, Christie et al. \cite{christie2020learning} proposed learning geocentric pose for rectifying oblique monocular satellite images and produced the first public dataset for this task, including satellite images over three cities in the United States. Their model (\figref{fig:model-diagrams}) relates height as a prior for vector field magnitude without an adequate mechanism to learn the true relationship, resulting in poor estimates for tall buildings. In our work, we explicitly model this relationship based on the geometry of affine projection, resulting in more accurate predictions for tall objects. We are currently not aware of any other published work on this topic. To promote further research and enable a public leaderboard for evaluation, we extend the dataset from \cite{christie2020learning} to include a challenging city outside the United States and to address inconsistencies in the original dataset (\secref{sec:datasets}).

\section{Affine Geocentric Pose}
\label{sec:affine-camera}
\begin{figure}[t!]
	\centering
	\includegraphics[width=\columnwidth]{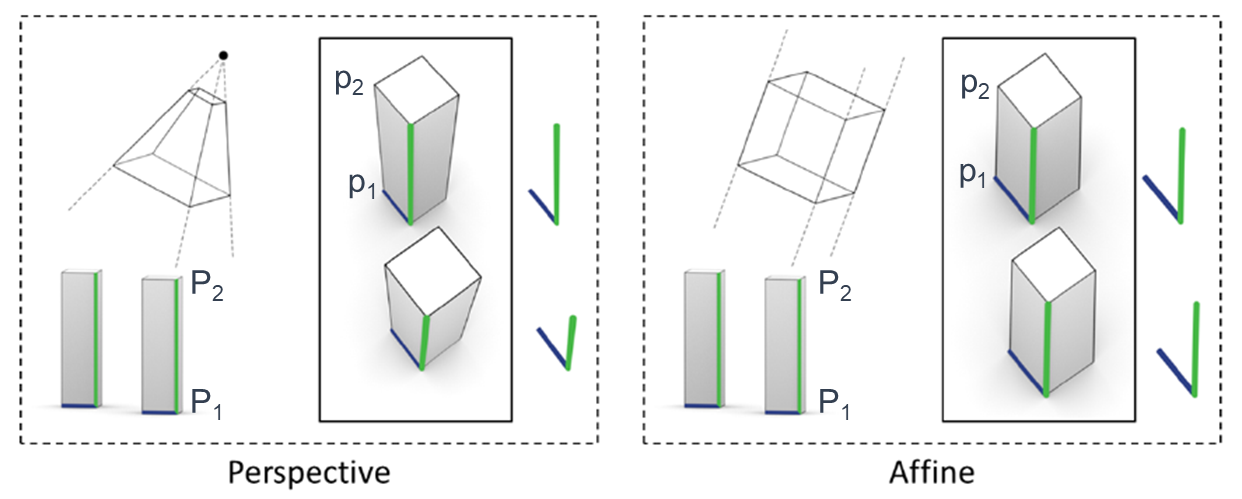}
	\caption{Affine projection $\boldsymbol{A}$ maps world coordinates $\boldsymbol{P}$ to image coordinates $\boldsymbol{p}$, preserving invariant properties of parallelism and ratio of lengths on parallel lines.}
	\label{fig:affine-geometry}
\end{figure}

Satellite pushbroom sensors are well-approximated locally (e.g., for processing image tiles) as affine cameras \cite{de2014stereo}, so we define our geocentric pose representation explicitly for an affine camera. Depth variations for objects in the scene are much smaller than depth from the sensor, angular field of view is narrow for a local sub-image, and the sensor maintains approximately the same attitude and speed throughout image acquisition. \figref{fig:affine-geometry} illustrates the invariant properties of affine projection compared to the more general perspective projection model, as described in detail in \cite{hartley2003multiple}. We exploit the invariant property of parallelism to define a single angle of parallel projection $\theta $ for each image. Unlike \cite{christie2020learning}, we also exploit preservation of the ratio of lengths on parallel lines to relate object heights with magnitudes of vectors mapping surface features to ground level.

We now define the mathematical assumptions in our model. Given image $\boldsymbol{I}$ with 2x3 affine projection matrix $\boldsymbol{A}$, we define geocentric pose $g\left(\boldsymbol{I}\right)=\left\{s,\theta ,\boldsymbol{h}\right\}$, where the vector $\boldsymbol{h}$ is height above ground level (AGL) for each pixel in image $\boldsymbol{I}$, $\theta $ is the angle of parallel projection in the image plane, and $s$ is the scale factor relating lengths of lines along those parallel projections in world coordinates and their corresponding lengths in image coordinates.

First we define affine projection $\boldsymbol{A}$ of any two vertically aligned world coordinates $P_1$ and $P_2$ to their image coordinates $p_1$ and $p_2$ observed in image $\boldsymbol{I}$.
\begin{align}
&p_1={\left( \begin{array}{cc}x_1 & y_1\end{array}\right)}^T=\boldsymbol{A}P_1 \\
&p_2={\left( \begin{array}{cc}x_2 & y_2 \end{array}\right)}^T=\boldsymbol{A}P_2 \\
&P_1={\left( \begin{array}{ccc}X & Y & Z_1 \end{array}\right)}^T \\
&P_2={\left( \begin{array}{ccc}X & Y & Z_2 \end{array}\right)}^T; Z_2>Z_1
\end{align}

We then define height $h$ as the distance between $P_1$ and $P_2$ (meters) and magnitude $m$ as the distance between $p_1$ and $p_2$ (pixels).
\begin{equation}
h=Z_2-Z_1; m=\left\|p_2-p_1\right\| 
\end{equation}
Given these two image coordinates $p_1$ and $p_2$, height $h$ (meters), and corresponding projected magnitude $m$ (pixels), we determine the angle of parallel projection $\theta $ (radians) and scale factor $s$ (pixels/meter). Observe that scale $s$ is zero for local nadir imaging geometry.
\begin{equation}\label{eq:scale-factor} 
\theta =atan2\left(y_2-y_1,x_2-x_1\right); s={m}/{h}
\end{equation}
Without loss of generality, we constrain all points ${\boldsymbol{P}}_{\boldsymbol{1}}$ to be at ground level and we define the vector field $\dot{\boldsymbol{p}}$ mapping surface features ${\boldsymbol{p}}_{\boldsymbol{2}}$ to ground level ${\boldsymbol{p}}_{\boldsymbol{1}}$ in the image plane.
\begin{equation}
\dot{\boldsymbol{p}}=\boldsymbol{m}\left( \begin{array}{c}
{\mathrm{cos} \theta \ } \\ 
{\mathrm{sin} \theta \ } \end{array}
\right)\boldsymbol{=}s\boldsymbol{h}\left( \begin{array}{c}
{\mathrm{cos} \theta \ } \\ 
{\mathrm{sin} \theta \ } \end{array}
\right)
\end{equation}

This representation for geocentric pose is valid for satellite images with locally affine imaging geometry; however, often satellite images are distributed in orthorectified form. Orthorectification for a single image is a pixel remapping that approximates orthographic, z-axis aligned, projection with fixed pixel scale (meters) and minimizes terrain relief displacement using a ground-level elevation model \cite{mcglone2004manual}. For an affine camera, we define orthorectification as the pixel remapping $w_Z$ acquired by inverting 2x3 affine matrix $\boldsymbol{A}$ with elevation model ${\boldsymbol{z}}_{\boldsymbol{0}}$ and scaling by ${\boldsymbol{K}={diag(k,k)}}$ defined by the desired pixel scale.
\begin{equation}
\arraycolsep=1.4pt
w_Z\left(\boldsymbol{p};{\boldsymbol{z}}_{\boldsymbol{0}}\right)=\boldsymbol{K}{
\left( \begin{array}{cc}
a_{11} & a_{12} \\ 
a_{21} & a_{22} \end{array}
\right)}^{-1}\left[\boldsymbol{p}\boldsymbol{-}\left( \begin{array}{c}
a_{13} \\ 
a_{23} \end{array}
\right){\boldsymbol{z}}_{\boldsymbol{0}}\right]
\end{equation}

This function is invertible given the elevation model and camera metadata. For scenes with little terrain relief such that ${\boldsymbol{z}}_{\boldsymbol{0}}$ is approximately a constant or linear function, the orthorectified projection is approximately affine with equivalent scale and angle values for estimating geocentric pose. We demonstrate only a modest reduction in accuracy for our model applied to orthorectified images (\tabref{tab:ortho-results}).

\section{Methods}

\subsection{Model and Optimization}
\label{sec:model}

\begin{figure}[ht!]
	\centering
	\includegraphics[width=\columnwidth]{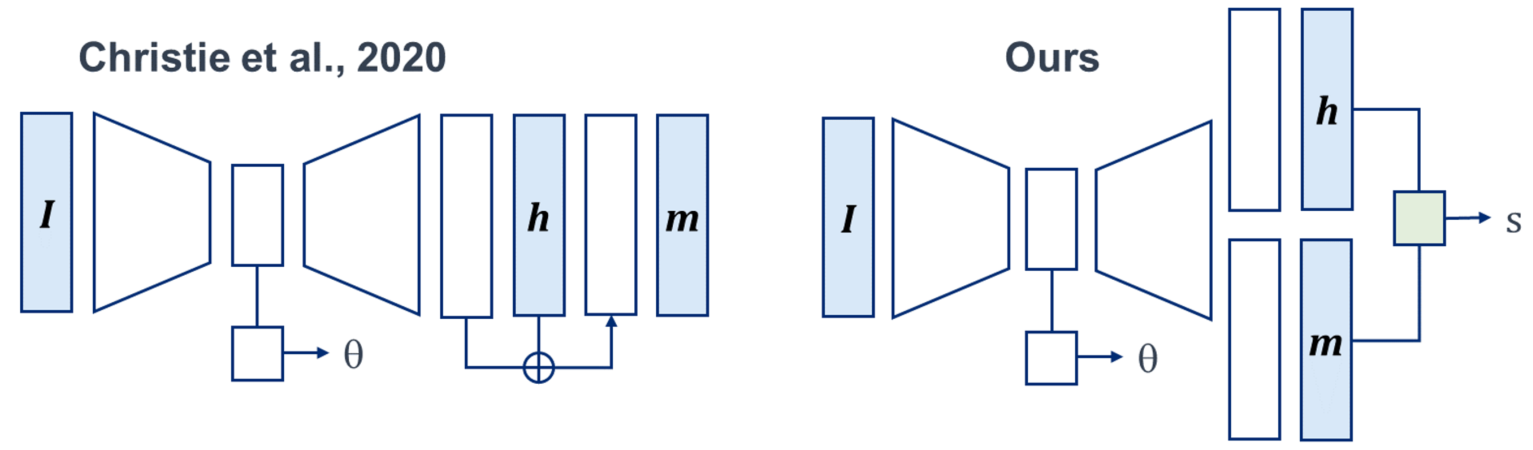}
	\caption{Christie et al. \cite{christie2020learning} relate height as a prior for vector magnitude. Our model explicitly encodes the known affine relationship with a custom least squares solver layer.}
	\label{fig:model-diagrams}
\end{figure}

We estimate geocentric pose $g\left(\boldsymbol{I}\right)=\left\{s,\theta ,\boldsymbol{h}\right\}$ with scale prediction defined by the known relationship between magnitude and height defined in \eqref{eq:scale-factor}. We parameterize our model in a multi-task deep network with a ResNet-34 encoder \cite{7780459} and U-Net decoder \cite{RFB15a} and supervise training with known height AGL values derived from lidar and multi-view stereo (MVS). We attach independent output heads for height and vector field magnitudes to the decoder, with consistency encouraged by solving for image-level scale as described below and shown in \figref{fig:model-diagrams}. This overcomes one of the most significant weaknesses of \cite{christie2020learning} which relates height as a prior for magnitude with an insufficient mechanism to learn the known relationship.

We represent vector fields with magnitude $\boldsymbol{m}$ and angle $\theta $ encoded as the vector $\left({\mathrm{cos} \theta },{\mathrm{sin} \theta }\right)$ to avoid ambiguity at zero. We explicitly supervise learning of both magnitude $\boldsymbol{m}$ and height $\boldsymbol{h}$ as orthogonal observations to enrich the features embedded in the encoder and decoder layers. We also explicitly supervise learning of scale which provides a gradient for learning to predict consistent magnitude and height values from images with or without explicit height labels. For some applications, scale will be available in metadata, so in addition to helping enforce consistency during training, having it as a prediction at test time can help identify regions where the model is performing poorly. Scale is predicted in the model with a custom least squares solver layer implemented using the pseudo-inverse, $s={\left({\boldsymbol{h}}^T\boldsymbol{h}\right)}^{-1}{\boldsymbol{h}}^T\boldsymbol{m}$.

The total loss $L$ minimized in training is a weighted sum of mean squared error (MSE) losses for all terms, 
${L=f_{\theta }L}_{\theta }{+f_sL}_s+{f_{\boldsymbol{h}}L}_{\boldsymbol{h}}+{f_{\boldsymbol{m}}L}_{\boldsymbol{m}}$. For height and magnitude, MSE is implemented as the mean of MSE values for each labeled image in a batch to both reduce sensitivity to unlabeled pixels and allow for training images without height labels to directly supervise height and magnitude; in the latter case, height and magnitude MSE losses are zero and only the scale loss is back-propagated through those layers. We set weighting factors $f_{\theta }$=10, $f_s$=10, $f_{\boldsymbol{h}}$=1, and $f_{\boldsymbol{m}}$=2 to normalize value ranges. To improve training efficiency and allow for a larger batch size, we down-sample input images by a factor of two. Our batch size b=8 speeds convergence for angle prediction since each training image provides only a single target sample. We use the Adam optimizer \cite{Kingma2015AdamAM} with a learning rate of 1e-4 which we found to improve convergence. For all experiments, we trained models for 200 epochs.

\subsection{Augmentation}
\label{sec:augmentation}
The distributions for angle of parallel projection $\theta$, scale factor $s$ relating height and magnitude, and object height $\boldsymbol{h}$ are all heavily biased. Angle and scale are biased by the limited viewing geometries from satellite orbits, and very tall objects are rare. We encourage generalization and address bias with image remap augmentations $w_{\theta }$, $w_s$, and $w_h$.
\begin{align}
&w_{\theta }\left(\boldsymbol{p};\dot{\theta }\right)=\left( \begin{array}{cc}
{\mathrm{cos} \dot{\theta }\ } & {\mathrm{sin} \dot{\theta }\ } \\ 
-{\mathrm{sin} \dot{\theta }\ } & {\mathrm{cos} \dot{\theta }\ } \end{array}
\right)\boldsymbol{p} \\
&w_s\left(\boldsymbol{p};\dot{s}\right)=\left( \begin{array}{cc}
\dot{s} & 0 \\ 
0 & \dot{s} \end{array}
\right)\boldsymbol{p} \\
&w_h\left(\boldsymbol{p};\dot{h}\right)=\boldsymbol{p}\boldsymbol{+}\dot{\boldsymbol{m}}\left( \begin{array}{c}
{\mathrm{cos} \theta \ } \\ 
{\mathrm{sin} \theta \ } \end{array}
\right) \\
&\dot{\boldsymbol{m}}\boldsymbol{=-}s\left(\boldsymbol{h}\boldsymbol{+}\dot{h}\right)
\end{align}

\begin{figure}[ht!]
	\centering
	\includegraphics[width=\columnwidth]{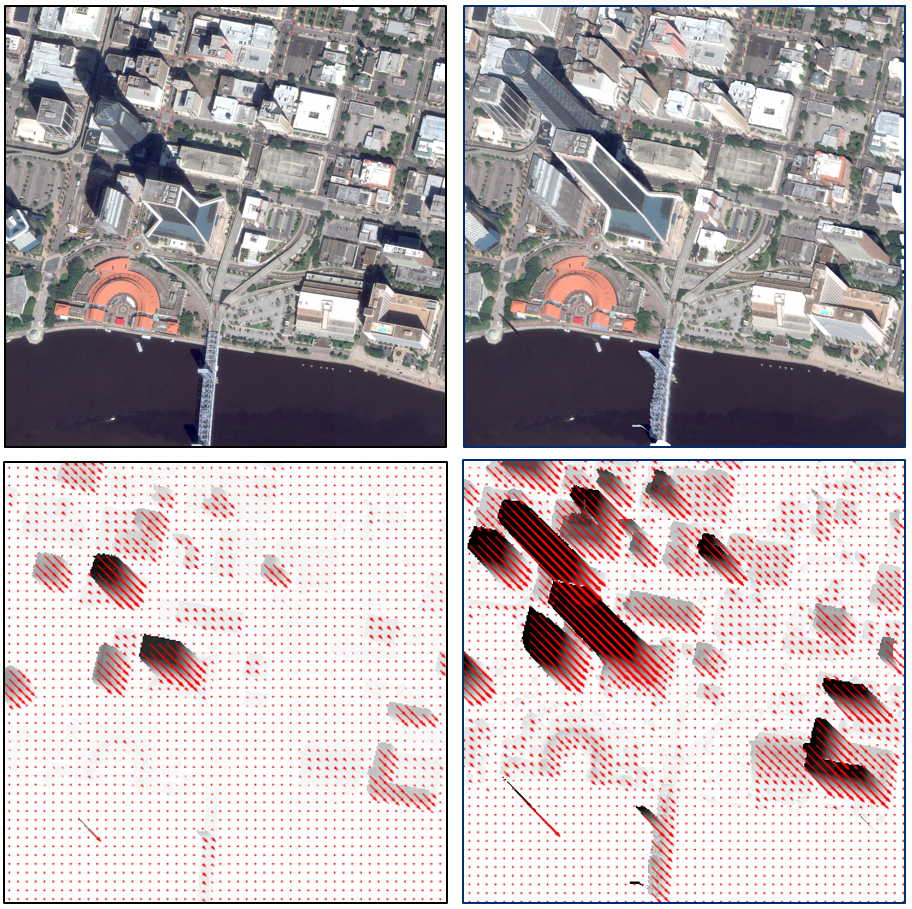}
	\caption{Affine geocentric pose enables height augmentations to address the long-tailed nature of height distributions in the data. For this example, we remap above-ground features in the original image (left) with 2.3x height (right) for RGB images (top) and geocentric pose values (bottom).}
	\label{fig:height-augmentations}
\end{figure}

While image augmentations for rotation $w_{\theta }$ and scale $w_s$ are commonly applied to regularize training in deep networks and depend only on image coordinates, our height augmentation $w_h$ inverts geocentric pose vectors to synthetically increase building heights (\figref{fig:height-augmentations}). Note that shadows are not adjusted by this simple but effective augmentation. This does not appear to be an impediment, and we believe that over-reliance on shadows should not be encouraged in learning because they are very often not observed.

\subsection{MVS Supervision}
\label{sec:supervision}
The geocentric pose method presented in \cite{christie2020learning} relies entirely on normalized digital surface model (nDSM) products derived from airborne lidar to define height above ground for supervising learning; however, lidar is not available or practical to collect in many world regions. MVS methods for satellite images \cite{s2p8014932,VisSat19,Shean2016AnAO, zcanli2015ACO, Qin2016RPCSP, Rupnik20183DRF} offer a more widely available alternative to lidar for training our model, and accuracy for these methods is comparable for larger features relevant to mapping applications \cite{bosch2016mvs3dm, Leotta_2019_CVPR_Workshops, j9229514}. To explore this, we adapted the Urban Semantic 3D (US3D) rendering pipeline \cite{bosch2019semantic} to produce labeled datasets using height derived from MVS. In our experiments, we demonstrate that models trained with lidar and commercially acquired MVS derived heights both perform well (\tabref{tab:vricon-results}).

\section{Experiments}
\label{sec:experiments}

\subsection{Datasets}
\label{sec:datasets}

We demonstrate our methods with the US3D public dataset \cite{bosch2019semantic}, first developed for the 2019 IEEE GRSS Data Fusion Contest (DFC19) \cite{le20192019} and later extended by Christie et al. \cite{christie2020learning} to explore the geocentric pose regression task. US3D includes satellite images and lidar-derived reference labels covering Jacksonville, Florida (JAX), Omaha, Nebraska (OMA), and Atlanta, Georgia (ATL). US3D images are 2048x2048 pixels except for those over ATL with off-nadir angle greater than 35 degrees. Those images were cropped much smaller, making them inconsistent with the rest of the dataset. Far off-nadir commercial satellite images are uncommon and often undesirable, as described by \cite{weir2019spacenet}. For better consistency, we excluded the cropped images for training and testing our models.

\begin{figure*}[t!]
	\includegraphics[width=\textwidth]{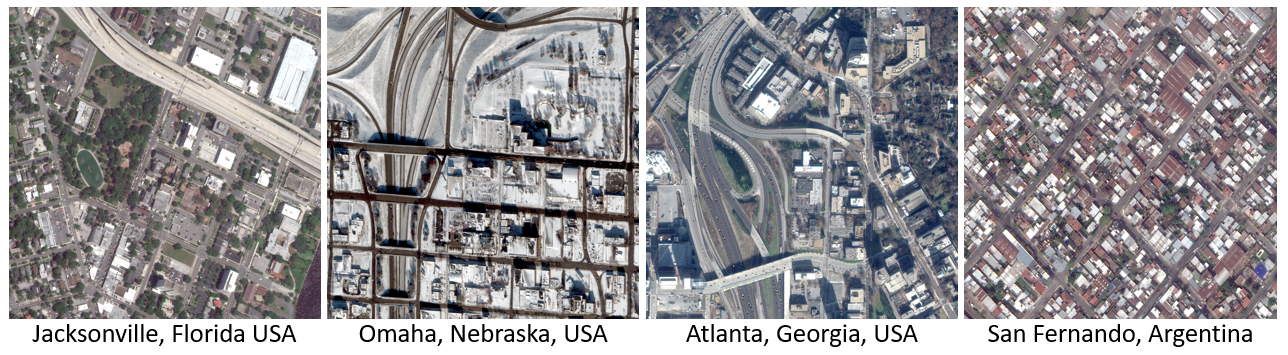}
	\caption{US3D includes satellite images from WorldView 2 and WorldView 3 and covers a variety of geographic location, season, viewpoint, and resolution. We added data for San Fernando, Argentina which presents new challenges, with fewer tall buildings and increased architectural diversity. Example images are shown. Statistics are provided in a supplement \cite{pubgeo_monocular_geocentric_pose}.}
	\label{fig:sites}
\end{figure*}

We further extended the public US3D dataset to include satellite images of San Fernando, Argentina (ARG) from the 2016 Multi-View Stereo 3D Mapping Challenge \cite{bosch2016mvs3dm}. Until now, US3D was limited to cities within the United States with Western architecture. ARG presents additional challenges, with fewer tall buildings and increased diversity in architectural styles, including very closely spaced buildings. Our current model does not perform as well for ARG as for the other sites, and we hope that publicly releasing this data will encourage further research to improve performance. Our extended dataset is illustrated in \figref{fig:sites}. Data statistics are provided in a supplement \cite{pubgeo_monocular_geocentric_pose}.

\subsection{Results}
\label{sec:results}
\begin{figure}[t!]
	\centering
	\includegraphics[width=0.75\columnwidth]{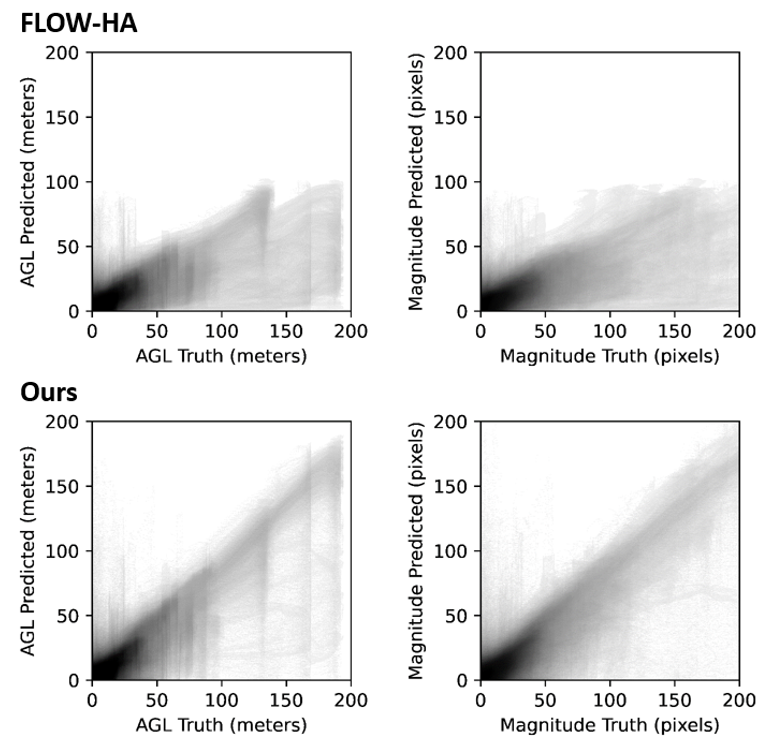}
	\caption{AGL heights and vector magnitudes are compared to reference values for the DFC19 test set. Pixel intensity indicates count. Our model significantly improves upon FLOW-HA \cite{christie2020learning}, particularly for infrequent tall buildings.}
	\label{fig:dfc19-accuracy-comparison}
\end{figure}

$\textbf{State of the Art for Geocentric Pose:}$ We first demonstrate that our model dramatically improves upon the state of the art method presented by Christie et al. \cite{christie2020learning} which, to the best of our knowledge, is the only published work on this task. Their FLOW-H model was trained without augmentations and FLOW-HA with rotation augmentations. We present results for our model trained with and without augmentations and trained with only the DFC19 train set including Jacksonville and Omaha, with only the ATL train set, and with the train sets from all four cities. Results are shown in \tabref{tab:cvpr-dfc19-comparison} and \figref{fig:dfc19-accuracy-comparison} for the DFC19 test set and in \tabref{tab:cvpr-atlsn4-comparison} for the ATL test set. We report root mean square error (RMSE) values for magnitude (pixels), angle (degrees), height (meters), and endpoint error (pixels). Mean absolute error (MAE) was reported by \cite{christie2020learning}, so we also demonstrate our improvements in terms of MAE in a supplement \cite{pubgeo_monocular_geocentric_pose}. Observe that our model design and augmentations each improve performance and reduce over-fitting for DFC19 but that augmentations sometimes harm performance for ATL. We believe this is because both train and test ATL images were drawn from a much less diverse single pass, same day satellite collection over a single city. Also note that training on additional cities consistently improves angle predictions for both test sets.

For both test sets, our model outperforms FLOW-HA even with no augmentations. We believe a primary contributing factor to \cite{christie2020learning} underestimating heights for taller buildings is its inadequate modeling of the relationship between height and vector field magnitude. As evidenced by our metric results and visually in \figref{fig:dfc19-accuracy-comparison}, our model that more directly relates height and magnitude with an image-level scale factor produces more accurate predictions.

To evaluate FLOW-HA performance for our new ARG test set, we trained it from scratch with all four cities and compared to our model both with and without augmentations (\tabref{tab:cvpr-arg-comparison}). Since very few buildings in this region are tall, FLOW-HA is more competitive with our model trained without augmentations; however, our model trained with augmentations provides a significant improvement.

\begin{table}[t!]
	\setlength{\tabcolsep}{5pt}
	\resizebox{\columnwidth}{!}{
		\begin{tabular}{llcccc}  
			\toprule  
			Method & Train & Mag & Angle & Endpoint & Height \\ 
			\midrule
			FLOW-HA \cite{christie2020learning} & DFC19 & 7.08 & 29.05 & 7.57 & 6.12 \\  
			FLOW-H \cite{christie2020learning} & DFC19 & 6.12 & 21.84 & 7.09 & 5.61 \\  
			Ours-NoAug & DFC19 & 5.06 & 16.53 & 5.58 & 4.81 \\  
			Ours & DFC19 & $\boldsymbol{4.07}$ & 13.73 & 4.31 & 4.00 \\  
			\midrule
			Ours-NoAug & All Cities & 5.19 & 20.51 & 5.91 & 4.86 \\  			
			Ours & All Cities & 4.14 & $\boldsymbol{11.79}$ & $\boldsymbol{4.29}$ & $\boldsymbol{3.89}$ \\   
			\bottomrule
		\end{tabular}
	}
    \
	\caption{Our method improves on state of the art RMSE for the DFC19 test set.}
	\label{tab:cvpr-dfc19-comparison}
\end{table}

\begin{table}[t!]
	\setlength{\tabcolsep}{5pt}
	\resizebox{\columnwidth}{!}{
		\begin{tabular}{llcccc}  
			\toprule  
			Method & Train & Mag & Angle & Endpoint & Height \\ 
			\midrule
			FLOW-HA \cite{christie2020learning} & ATL & 7.53 & 20.54 & 8.19 & 10.58 \\  
			FLOW-H \cite{christie2020learning} & ATL & 5.62 & 14.41 & 5.88 & 8.95 \\  
			Ours-NoAug & ATL & 3.50 & 11.00 & 3.93 & 4.88 \\  
			Ours & ATL & $\boldsymbol{3.45}$ &  13.19 & 3.88 & 4.89 \\  
			\midrule
			Ours-NoAug & All Cities & 3.46 & $\boldsymbol{10.67}$ & 3.83 & $\boldsymbol{4.84}$ \\  			
			Ours & All Cities & 3.50 & $\boldsymbol{10.67}$ & $\boldsymbol{3.72}$ & 4.95 \\   
			\bottomrule
		\end{tabular}
	}
    \
	\caption{Our method improves on state of the art RMSE for the ATL test set.}
	\label{tab:cvpr-atlsn4-comparison}
\end{table}

\begin{table}[t!]
	\centering
	\setlength{\tabcolsep}{5pt}
	\resizebox{0.83\columnwidth}{!}{
		\begin{tabular}{lccccc}  
			\toprule  
			Method & Mag & Angle & Endpoint & Height \\ 
			\midrule
			FLOW-HA \cite{christie2020learning} & 3.82 & 31.96 & 4.21 & 3.09 \\ 
			Ours-NoAug & 3.97 & $\boldsymbol{19.89}$ & 4.14 & 3.52 \\     
			Ours & $\boldsymbol{3.32}$ & 23.28 & $\boldsymbol{3.56}$ & $\boldsymbol{3.00}$ \\  			
			\bottomrule
		\end{tabular}
	}
	\
	\caption{Our method improves on state of the art RMSE for the challenging new ARG test site. All models were trained on all four sites.}
	\label{tab:cvpr-arg-comparison}
\end{table}

$\textbf{State of the Art for Height Prediction:}$ We also compare our results with methods for pixel-level height prediction using both our full geocentric pose model and the same model with only a height prediction head. For fair comparison with \cite{mahmud2020boundary,mou2018im2height,srivastava2017joint,christie2020learning}, we retrained and tested our models and \cite{christie2020learning} using the custom DFC19 train and test sets used in \cite{mahmud2020boundary}. For comparison with \cite{kunwar2019u,zheng2019pop,christie2020learning}, we tested against the DFC19 test set used for \cite{j9229514}. Results are reported in \tabref{tab:height-comparison} and \tabref{tab:height-comparison-two}, respectively. Both models achieve state of the art accuracy, with our single-task model  outperforming the multi-task model. We attribute this improvement to our augmentations that address biases in the data.

\begin{table}[t!]
	\centering
	\setlength{\tabcolsep}{5pt}
	\resizebox{0.63\columnwidth}{!}{
		\begin{tabular}{lcc}
			\toprule  
			Method & MAE & RMSE \\ 
			\midrule
			Srivastava et al. \cite{srivastava2017joint} & 3.74 & 5.85 \\ 
			Mou and Zhu \cite{mou2018im2height} & 3.62 & 5.40 \\     
			Mahmud et al. \cite{mahmud2020boundary} & 3.34 & 5.02 \\
			Christie et al. \cite{christie2020learning} & 2.76 & 4.33 \\
			Ours (Height) & $\boldsymbol{2.52}$ & $\boldsymbol{3.89}$ \\
			Ours (Full) & 2.78 & 4.07 \\
			\bottomrule
		\end{tabular}
	}
	\
	\caption{Our pixel-level building height predictions improve on state of the art MAE and RMSE (meters) for the custom DFC19 test set used by \cite{mahmud2020boundary}.}
	\label{tab:height-comparison}
\end{table}

\begin{table}[t!]
	\centering
	\setlength{\tabcolsep}{5pt}
	\resizebox{\columnwidth}{!}{
		\begin{tabular}{lcccc}
			\toprule  
			Method & All MAE & All RMSE & Bldgs MAE & Bldgs RMSE\\ 
			\midrule
			Kunwar \cite{kunwar2019u} & 2.69 & 9.26 & 8.33 & 19.65 \\
			Zheng et al. \cite{zheng2019pop} & 2.94 & 9.24 & 8.72 & 19.32 \\	
			Christie et al. \cite{christie2020learning} & 2.98 & 8.23 & 7.73 & 16.87 \\	
			Ours (Height) & $\boldsymbol{2.31}$ & $\boldsymbol{5.46}$ & $\boldsymbol{5.75}$ & $\boldsymbol{10.69}$ \\
			Ours (Full) & 2.44 & 5.76 & 6.14 & 11.35 \\	
			\bottomrule
		\end{tabular}
	}
	\
	\caption{Our pixel-level height predictions improve on state of the art MAE and RMSE (meters) for the DFC19 test set used for \cite{j9229514}.}
	\label{tab:height-comparison-two}
\end{table}

$\textbf{Geographic Diversity:}$ We assess model performance for geographically diverse cities shown in \figref{fig:sites}. Results are shown in \tabref{tab:city-results} for our full model, the same model trained without augmentations, a version with only a height regression head, and our model fine tuned for twenty epochs on each individual city. We report RMSE for angle, scale, height, and vector field endpoints. Here we see a modest improvement in height prediction with our full model, though not for ATL which includes only images from a single pass as discussed above.

We might expect fine tuning our model for each site to notably improve performance given unique geographic appearance and differences in the image resolutions among sites. After fine tuning, we see similar or modestly improved performance for all sites except for OMA which has limited appearance diversity.

In a supplement \cite{pubgeo_monocular_geocentric_pose}, we also report the $R^2$ metric for height and endpoint and discuss its value for comparing performance among cities. Height prediction is especially challenging for ARG due to both closely spaced residential housing and the infrequency and relative uniqueness of the taller buildings. While RMSE is low relative to other sites due to the predominance of low height buildings, the $R^2$ metric, which is normalized by the value range, is poor due to inaccurate predictions for rare tall structures.

\begin{table}[t!]
    \centering
	\setlength{\tabcolsep}{5pt}
	\resizebox{\columnwidth}{!}{
		\begin{tabular}{llcccc}  
			\toprule  
			Prediction & City & Full & NoAug & Height & Tuned \\ 
			\midrule
			Angle (deg) & JAX & 12.48 & 20.40 & N/A & $\boldsymbol{10.26}$ \\ 
			& OMA & $\boldsymbol{11.31}$ & 20.58 & N/A & 17.90 \\ 
			& ATL & 10.67 & 10.67 & N/A & $\boldsymbol{9.89}$ \\ 			
			& ARG & 23.28 & $\boldsymbol{19.89}$ & N/A & 20.48 \\
			\midrule
			Scale (pix/m) & JAX & $\boldsymbol{0.11}$ & 0.17 & N/A & $\boldsymbol{0.11}$ \\ 
			& OMA & $\boldsymbol{0.11}$ & 0.16 & N/A & $\boldsymbol{0.11}$ \\ 
			& ATL & 0.10 & 0.12 & N/A & $\boldsymbol{0.08}$ \\ 			
			& ARG & 0.16 & 0.19 & N/A & $\boldsymbol{0.13}$ \\
			\midrule
			Height (m) & JAX & 3.33 & 3.69 & 3.67 & $\boldsymbol{3.23}$ \\ 
			& OMA & $\boldsymbol{4.15}$ & 5.51 & 4.20 & 4.32 \\ 
			& ATL & 4.86 & 4.84 & 4.73 & $\boldsymbol{4.67}$ \\ 			
			& ARG & $\boldsymbol{3.00}$ & 3.52 & 3.27 & 3.06 \\
			\midrule
			Endpoint (pix) & JAX & 3.61 & 3.69 & N/A & $\boldsymbol{3.53}$ \\ 
			& OMA & $\boldsymbol{4.63}$ & 6.80 & N/A & 5.19 \\ 
			& ATL & 3.66 & 3.83 & N/A & $\boldsymbol{3.45}$ \\ 			
			& ARG & $\boldsymbol{3.56}$ & 4.14 & N/A & 3.58 \\
			\bottomrule
		\end{tabular}
	}
	\
	\caption{RMSE is compared with our full model, without augmentations, with only a height prediction head, and with the full model fine tuned for each city.}
	\label{tab:city-results}
\end{table}

$\textbf{Test-time Augmentations:}$ To better understand our model's ability to generalize, we tested models trained with and without augmentations on all cities with random rotation, scale, and height augmentations. Results in \tabref{tab:test-augmentations} show significant improvement in generalization to account for a broader range of conditions than observed in the limited test sets. Our height augmentations help overcome the issue of long-tailed height distributions in the training data by significantly improving height estimation both with and without test-time height augmentations; however, it does not learn to generalize for these appearance changes as well as for scale and rotation angle. Also note that angle predictions are more accurate for images with test-time height augmentations because angles are more observable when vector magnitudes are larger.

\begin{table}[t!]
    \centering
	\setlength{\tabcolsep}{5pt}
	\resizebox{0.76\columnwidth}{!}{
		\begin{tabular}{llccc}  
			\toprule  
			Prediction & Test Aug & Ours & NoAug \\ 
			\midrule
			Angle (deg) & None & 17.73 & $\boldsymbol{17.72}$ \\
			& Scale & $\boldsymbol{19.00}$ & 34.93 \\
			& Angle & $\boldsymbol{20.18}$ & 87.72 \\
			& Height & $\boldsymbol{14.61}$ & 17.92 \\
			\midrule
			Scale (pix/m) & None & $\boldsymbol{0.13}$ & 0.17 \\
			& Scale & $\boldsymbol{0.18}$ & 0.28 \\
			& Angle & $\boldsymbol{0.15}$ & 0.29 \\
			& Height & $\boldsymbol{0.17}$ & $\boldsymbol{0.17}$ \\
			\midrule
			Height (m) & None & $\boldsymbol{3.84}$ & 4.67 \\
			& Scale & $\boldsymbol{4.24}$ & 5.34 \\
			& Angle & $\boldsymbol{4.20}$ & 8.17 \\
			& Height & $\boldsymbol{6.18}$ & 7.27 \\
			\midrule
			Endpoint (pix) & None & $\boldsymbol{3.84}$ & 4.84 \\
			& Scale & $\boldsymbol{5.05}$ & 7.17 \\			
			& Angle & $\boldsymbol{4.17}$ & 9.76 \\				
			& Height & $\boldsymbol{5.47}$ & 7.07 \\
			\bottomrule
		\end{tabular}
	}
	\
	\caption{Test-time augmentations demonstrate improved RMSE over a range of conditions not observed in the test set. Results are for test images from all four cities. \\}
	\label{tab:test-augmentations}
\end{table}

$\textbf{Orthorectified Images:}$ As discussed in \secref{sec:affine-camera}, satellite images are often orthorectified for convenient dissemination and georeferencing on a Cartesian grid. This orthorectification distorts images such that our affine assumptions are violated. To evaluate the impact on model performance, we applied orthorectification to all test images and compared accuracies to the original test images as shown in \tabref{tab:ortho-results}. Angle prediction is improved due to a smaller value range, but scale accuracy is degraded significantly. We then orthorectified the training images and fine tuned the model for twenty epochs. The resulting predictions are only modestly less accurate than their affine counterparts, indicating that our model can produce accurate predictions when applied to typical orthorectified images. Rare high terrain slopes (meters) are not present in our data set but can induce more significant rectification errors (pixels) in orthorectified images. We believe modest changes to our model implementation can address this issue, and we plan to pursue this in future work.

$\textbf{Building Segmentation:}$ We compare performance with non-building above-ground feature heights set to zero in train and test sets (\tabref{tab:ortho-results}). Performance is comparable for this narrowly-defined task, suggesting that our method can be used for building segmentation and rectification when semantic labels are available for training.

\begin{table}[t!]
	\setlength{\tabcolsep}{5pt}
	\resizebox{\columnwidth}{!}{
		\begin{tabular}{llcccc}  
			\toprule  
			Test Set & Tuned & Angle & Scale & Height & Endpoint \\ 
			\midrule
   			All pixels & No & 17.73 & $\textbf{0.13}$ & $\textbf{3.84}$ & $\textbf{3.84}$ \\  
   			All pixels ortho & No & 12.99 & 0.26 & 4.32 & 5.41 \\  
			All pixels ortho & Yes & $\textbf{10.65}$ & 0.19 & 3.95 & 4.57 \\  
			\midrule
   			Building pixels & No & 17.81 & 0.14 & $\textbf{3.48}$ & $\textbf{3.41}$ \\  
   			Buildings ortho & No & $\textbf{10.00}$ & 0.19 & 4.78 & 5.42 \\  
   			Buildings ortho & Yes & 10.70 & $\textbf{0.13}$ & 3.94 & 4.17 \\  
			\bottomrule
		\end{tabular}
	}
	\
	\caption{Predictions for orthorectified images where affine assumptions are violated are only modestly less accurate than their affine counterparts after fine tuning. Metrics shown are RMSE. All pixels results are for all four cities. Results for only building pixels do not include ARG.}
	\
	\label{tab:ortho-results}
\end{table}

\begin{figure}[h!]
	\includegraphics[width=0.86\columnwidth]{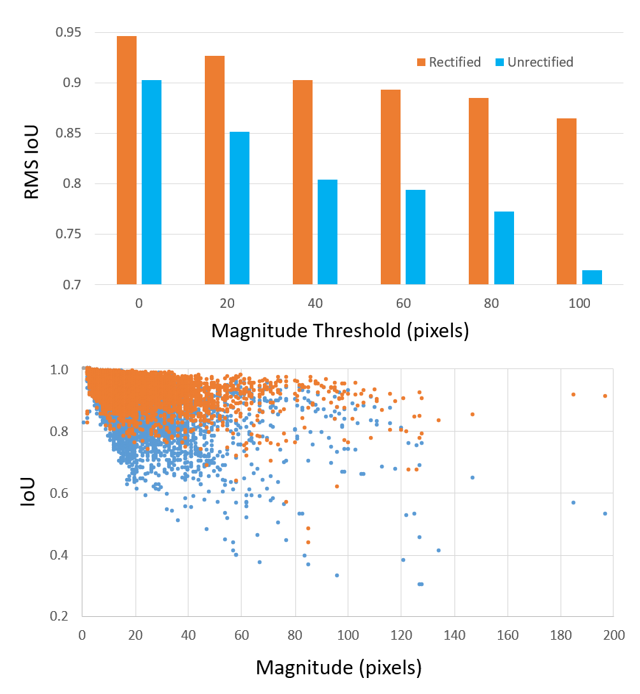}
	\caption{Instance-level building IoU improvement with rectification is shown versus max vector magnitude inside the instance (bottom), and RMS IoU is shown for instances with max magnitude above a threshold (top). \\}
	\label{fig:iou-comparison}
\end{figure}

$\textbf{Rectification to Ground Level:}$ Rectifying building segmentation labels into footprints is one of the applications of our work. We demonstrate the ability of our model to perform this task by plotting instance-level intersection over union (IoU) as a function of maximum vector magnitude inside each instance for both unrectified building labels and our rectified outputs compared to building footprints. We also show RMS IoU as a function of a maximum magnitude threshold for building instances included in the calculation (\figref{fig:iou-comparison}). In both cases, we evaluate a subset of the building test instances where warping with reference geocentric pose values achieves a minimum IoU of 0.9. This helps eliminate instances where occlusion prevents accurate geocentric pose regression.

$\textbf{MVS Supervision:}$ For experiments to supervise learning with MVS instead of lidar, discussed in \secref{sec:supervision}, we used commercial Vricon MVS products that overlap the DFC19 Jacksonville and Omaha train and test sets. While our current MVS dataset is limited to a subset of the DFC19 images (1461/225 train/test images), MVS data can be produced over large scales anywhere in the world from satellite images. We believe this is important for providing sufficient diversity in training to promote generalization.

We trained our model separately with lidar and MVS and then tested against both lidar and MVS reference values. As shown in \tabref{tab:vricon-results}, both models achieved comparable accuracies for both heights and vector field endpoints across both evaluation types. Models trained using one source of reference data performed better when evaluating against the same source (e.g., train and test with lidar), but both models performed well. Angle and scale predictions appear surprisingly less accurate for the model trained with MVS; however, these are image-level predictions, so the differences for only 225 test images are unlikely to be significant.

\begin{table}[t!]
    \centering
	\setlength{\tabcolsep}{5pt}
	\resizebox{0.9\columnwidth}{!}{
		\begin{tabular}{llcccc}  
			\toprule  
			Train & Test & Angle & Scale & Height & Endpoint \\ 
			\midrule			
			Lidar & Lidar & $\textbf{13.72}$ & $\textbf{0.13}$ & 4.56 & 5.01 \\
			Lidar & MVS & $\textbf{13.72}$ & $\textbf{0.13}$ & 5.49 & 5.96 \\
			MVS & Lidar & 19.23 & 0.15 & 5.06 & 5.64 \\
			MVS & MVS & 19.23 & 0.15 & $\textbf{4.52}$ & $\textbf{4.94}$ \\ 
			\bottomrule
		\end{tabular}
	}
	\
	\caption{RMSE for models supervised with MVS are comparable to those supervised with lidar.}
	\
	\label{tab:vricon-results}
\end{table}

\section{Discussion}
\label{sec:discussion}

Geocentric pose regression from oblique monocular remote sensing images has the potential to dramatically improve the utility of oblique images for mapping applications. In this work, we presented our method that exploits invariant properties of affine imaging geometry to achieve state of the art performance for this task and also for pixel-level height prediction. Our affine approximation is very accurate for the 2048 x 2048 pixel sub-images we extract from large satellite images. For efficient processing of full images, our method can be applied in parallel to overlapping image tiles.

We also addressed challenges that must be overcome to deploy a solution in the real world, including supervising learning without lidar, reducing bias in training to enable generalization, and ensuring good performance for orthorectified images which are commonly available but which violate affine assumptions. Results rectifying images, building labels, and heights to ground level indicate the value of our approach for a range of mapping tasks. 

Our open source code provides a strong baseline for public leaderboard evaluation. In a supplement \cite{pubgeo_monocular_geocentric_pose}, we provide additional details of the public dataset \cite{9frn-7208-20} and metrics for evaluation as well as examples of common failure cases to motive further research.

\section*{Acknowledgements}
This work was supported by the National Geospatial-Intelligence Agency and approved for public release, 21-483, with distribution statement A – approved for public release; distribution is unlimited. Commercial satellite images were provided courtesy of DigitalGlobe.

{\small
\bibliographystyle{ieee_fullname}
\bibliography{references}
}

\end{document}


\newcommand{\equationref}[1]{Equation~\ref{#1}}

\newcommand{\reducecaptionspace}{\vspace{-0.3cm}\xspace}
\newcommand{\reducepostfigspace}{\vspace{-0.3cm}\xspace}
\newcommand{\reducespace}{\vspace{-0.3cm}\xspace}

\newcommand{\atl}{\textsc{ATL-SN4}\xspace}
\newcommand{\dfc}{\textsc{DFC19}\xspace}
\newcommand{\sanfernando}{\textsc{ARG}\xspace}
\newcommand{\ner}{\textsc{NER}\xspace}

\newcommand{\rmse}{$RMSE$\xspace}
\newcommand{\rsquared}{$R^2$\xspace}
\newcommand{\tss}{$TSS$\xspace}
\newcommand{\rss}{$RSS$\xspace}

\newcommand{\todo}[1]{\textcolor{red}{TODO: #1}}

\newcommand{\flow}{\textsc{flow}\xspace}
\newcommand{\flowh}{\textsc{flow-h}\xspace}
\newcommand{\flowa}{\textsc{flow-a}\xspace}
\newcommand{\flowha}{\textsc{flow-ha}\xspace}

\newcommand{\specialcell}[2][c]{%
\begin{tabular}[#1]{@{}c@{}}#2\end{tabular}}

\newcommand{\subsubheaderbf}[1]{\mbox{\textbf{#1}\hspace*{2.5mm}}}

\title{Supplementary Material: Single View Geocentric Pose in the Wild}

\author{Gordon Christie$^{1}$, Kevin Foster$^1$, Shea Hagstrom$^1$, Gregory D. Hager$^2$, Myron Z. Brown$^1$ \\ 
$^1$The Johns Hopkins University Applied Physics Laboratory \\ 
$^2$Department of Computer Science, The Johns Hopkins University }

\maketitle
\thispagestyle{empty}

\noindent{In this supplementary document we provide:}
\begin{compactitem}
\item[] \ref{sec:data-stats}: Summary and discussion of data statistics
\item[] \ref{sec:metrics}: Expanded discussion of metrics and results
\item[] \ref{sec:limfacs}: Examples and discussion of limiting factors
\end{compactitem}

\section{Data Statistics}
\label{sec:data-stats}

Statistics for the datasets used in our experiments are summarized in \tabref{tab:data-stats}. Statistics for the public Urban Semantic 3D (US3D) dataset \cite{bosch2019semantic}, including terrain variation and height distributions, are separately reported by \cite{christie2020learning}. We extended US3D with new public data for San Fernando, Argentina which presents additional challenges, with fewer tall buildings and increased architectural diversity.

\figref{fig:scale_distributions} characterizes distributions of scale factor values that relate heights above ground to their respective vector field magnitudes for mapping surface-level features to ground level. Values are higher for more oblique images and close to zero for near-nadir viewing geometry. The train and test sets are well balanced.

Distributions for height above ground values are shown in \figref{fig:agl_distributions}. Our new train and test set for Argentina is well balanced, as are those from Atlanta and Omaha. While the overall DFC19 dataset \cite{j9229514} including both Jacksonville and Omaha are well balanced, the Jacksonville test set does not capture the full range of values represented in its train set.

\begin{table*}[t!]
	\caption{Statistics for our train and test sets.}	
	\resizebox{\textwidth}{!}{
		\begin{tabular}{lcccc}  
			\toprule  
			& Jacksonville, Florida (JAX) & Omaha, Nebraska (OMA) & Atlanta, Georgia (ATL) & San Fernando, Argentina (ARG) \\ 
			\midrule
			Train image chips & 1098 & 1796 & 704 & 2325 \\  
			Test image chips & 120 & 178 & 264 & 463 \\  
			Source satellite images & 24 & 43 & 30 & 39 \\  
			Train geographic tiles & 52 & 53 & 52 & 63 \\  
			Test geographic tiles & 5 & 5 & 22 & 13 \\  
			Imaging satellite & WorldView-3 & WorldView-3 & WorldView-2 & WorldView-3 \\  
			Pixel size range (cm) & 31 -- 39 & 31 -- 36 & 47 -- 59 & 31 -- 41 \\  
			Azimuth angle range (deg) & 3 -- 268 & 3 -- 349 & 0 -- 358 & 14 -- 352 \\  
			Elevation angle range (deg) & 57 -- 84 & 63 -- 86 & 56 -- 81 & 54 -- 84 \\  
			Year range & 2014 -- 2016 & 2014 -- 2015 & 2009 & 2015 \\  
			Max height above ground (m) & 200 & 200 & 200 & 100 \\  
			\bottomrule
		\end{tabular}
	}
	\label{tab:data-stats}
\end{table*}

\section{Metrics}
\label{sec:metrics}
In our paper, for consistency we report accuracy with root mean square error (RMSE). Results by \cite{christie2020learning} were reported as mean absolute error (MAE), so for completeness we demonstrate our improvements in terms of MAE in \tabref{tab:cvpr-dfc19-comparison-mae} and \tabref{tab:cvpr-atlsn4-comparison-mae}. There are small differences between our numbers and those reported in \cite{christie2020learning} because of minor dataset changes they made before public release.

\begin{figure}[h!]
    \centering
	\includegraphics[width=0.9\columnwidth]{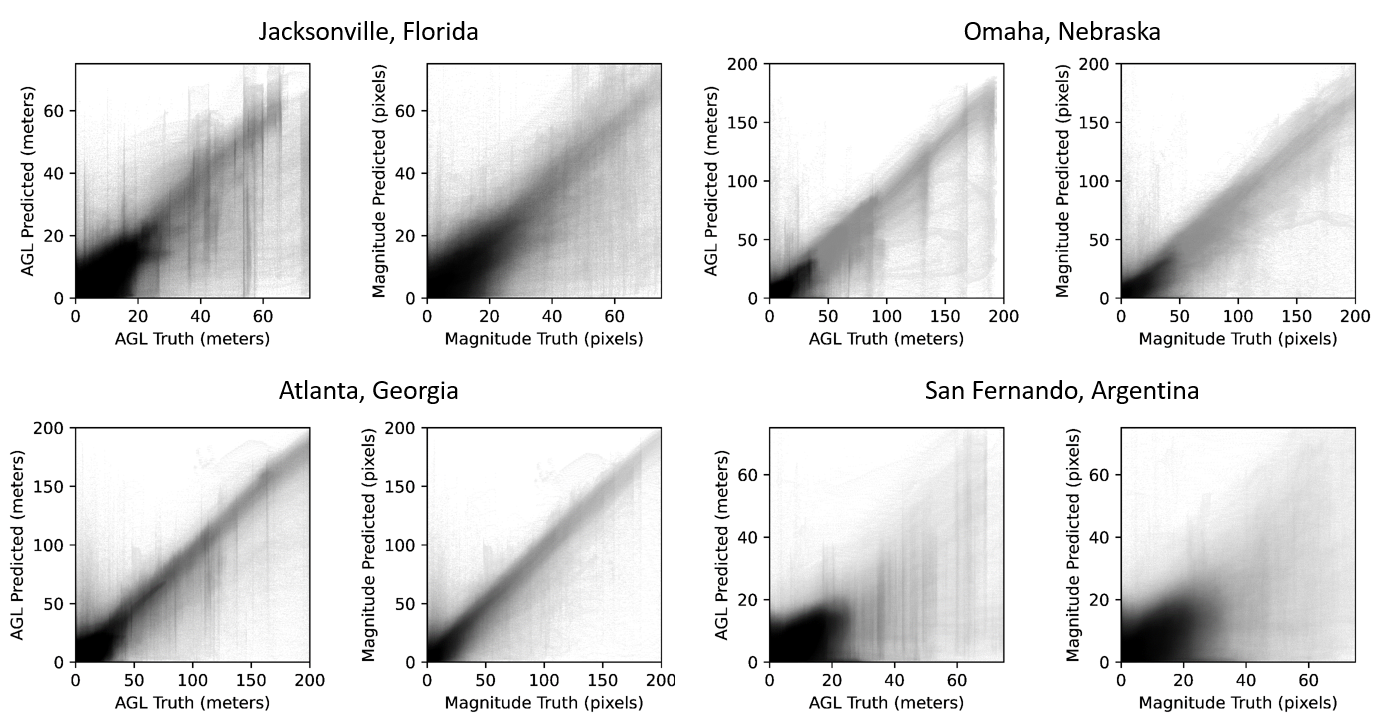}
	\caption{Above ground level (AGL) heights and vector field magnitudes from our model are compared to reference values for all test cities. Pixel intensity indicates count.}
	\label{fig:ours-scatter}
\end{figure}

For relative assessment of performance for multiple cities, we adopt the $R^2$ metric defined below and report results in \tabref{tab:city-results-r2}. $R^2$ clearly indicates relative prediction accuracy among cities, as shown in \figref{fig:ours-scatter}. In particular, $R^2$ correctly indicates that the predictive power of our regression model for ARG is much lower than for the other sites.

\begin{figure}[h!]
    \centering
	\includegraphics[width=0.8\columnwidth]{figs/scale_distributions.png}
	\caption{Histograms show distribution of image-level scale factors (pixels/meter) relating heights above ground to their respective vector field magnitudes for mapping surface pixels to ground level. Higher values represent images with more oblique viewing angle.}
	\label{fig:scale_distributions}
\end{figure}

\begin{figure}[h!]
    \centering
	\includegraphics[width=0.8\columnwidth]{figs/agl_distributions.png}
	\caption{Plots show the distributions of height above ground (meters) with $log10$ pixel counts for all sites.}
	\label{fig:agl_distributions}
\end{figure}

\begin{table}[h!]
	\centering
	\setlength{\tabcolsep}{5pt}
	\resizebox{0.9\columnwidth}{!}{
		\begin{tabular}{llcccc}  
			\toprule  
			Method & Train & Mag & Angle & Endpoint & Height \\ 
			\midrule
			FLOW-HA \cite{christie2020learning} & DFC19 & 2.62 & 16.82 & 3.00 & 2.26 \\  
			FLOW-H \cite{christie2020learning} & DFC19 & 2.32 & 15.58 & 2.80 & 2.14 \\  
			Ours-NoAug & DFC19 & 1.81 &  11.62 & 2.18 & 1.66 \\  
			Ours & DFC19 & $\boldsymbol{1.71}$ & 9.08 & 1.93 & 1.66 \\  
			\midrule
			Ours-NoAug & All Cities & 1.84 & 14.59 & 2.32 & 1.71 \\  			
			Ours & All Cities & 1.72 & $\boldsymbol{8.24}$ & $\boldsymbol{1.88}$ & $\boldsymbol{1.64}$ \\   
			\bottomrule
		\end{tabular}
	}
    \
	\caption{Our method improves on state of the art MAE errors for the DFC19 test set.}
	\label{tab:cvpr-dfc19-comparison-mae}
\end{table}

\begin{table}[h!]
	\centering
	\setlength{\tabcolsep}{5pt}
	\resizebox{0.9\columnwidth}{!}{
		\begin{tabular}{llcccc}  
			\toprule  
			Method & Train & Mag & Angle & Endpoint & Height \\ 
			\midrule
			FLOW-HA \cite{christie2020learning} & ATL & 3.53 & 15.54 & 4.12 & 4.68 \\  
			FLOW-H \cite{christie2020learning} & ATL & 2.79 & 9.27 & 3.05 & 4.00 \\  
			Ours-NoAug & ATL & $\boldsymbol{1.78}$ & 7.77 & 2.13 & $\boldsymbol{2.48}$ \\  
			Ours & ATL & 1.87 & 9.50 & 2.23 & 2.72 \\  
			\midrule
			Ours-NoAug & All Cities & 1.81 & 9.02 & 2.16 & 2.52 \\  			
			Ours & All Cities & 1.89 & $\boldsymbol{7.15}$ & $\boldsymbol{2.12}$ & 2.75 \\   
			\bottomrule
		\end{tabular}
	}
    \
	\caption{Our method improves on state of the art MAE errors for the ATL test set.}
	\label{tab:cvpr-atlsn4-comparison-mae}
\end{table}

\begin{table}[h!]
	\centering
	\setlength{\tabcolsep}{5pt}
	\resizebox{0.83\columnwidth}{!}{
		\begin{tabular}{lcccc}  
			\toprule  
			& JAX & OMA & ATL & ARG \\ 
			\midrule
			Height RMSE (m) & 3.33 & 4.15 & 4.86 & 3.00 \\
			Endpoint RMSE (pix) & 3.61 & 4.63 & 3.66 & 3.56 \\
			\midrule
			Height $R^2$ & 0.81 & 0.87 & 0.89 & 0.60 \\
			Endpoint $R^2$ & 0.84 & 0.88 & 0.90 & 0.68 \\
			\bottomrule
		\end{tabular}
	}
	\
	\caption{Our height and vector field prediction RMSE and $R^2$ are shown for four cities with significantly different value ranges. For RMSE, lower is better. Higher is better for $R^2$ $\in$ [0,1]. $R^2$ much more clearly indicates relative prediction accuracy among cities.}
	\label{tab:city-results-r2}
\end{table}

We define $R^2$ in terms of the residual sum of squares (\rss) of predicted values $f(x_i)$  for $n$ observed samples $x_i$ and reference values $y_i$ in \equationref{eq:rss}. RMSE, $\sqrt{RSS/n}$, is useful for measuring accuracy in units of the dependent variable $y_i$ (e.g., meters for heights and pixels for the vector field) for a single dataset; however, for multiple datasets with varying value scales (e.g., large building height values in urban scenes and smaller values in suburban scenes), a normalized metric is more discriminating for measuring the estimator’s predictive power. We normalize $RSS$ by the total sum of squares ($TSS$) of the dependent variable in \equationref{eq:tss}, leading to the coefficient of determination \rsquared in \equationref{eq:r2}. 

\begin{equation} 
RSS=\sum^n_{i=1}{{\left(y_i-f\left(x_i\right)\right)}^2} 
\label{eq:rss} 
\end{equation} 
\begin{equation} \label{eq:tss} 
TSS=\sum^n_{i=1}{{\left(y_i-\overline{y}\right)}^2} 
\end{equation} 
\begin{equation} \label{eq:r2} 
R^2=\max{(0,\ 1-\frac{RSS}{TSS})}
\end{equation} 
	
While $R^2$ is commonly applied for linear regression of trend lines, the general form measures the fraction of the total variance explained by any estimator’s predictions. Since $RSS$ can exceed $TSS$ for a poor prediction, we clip negative values to zero such that $R^2 \in$ [0,1].

\section{Examples and Limiting Factors}
\label{sec:limfacs}

$\textbf{State of the art performance:}$ Our method  exploits invariant properties of affine imaging geometry to achieve
state of the art performance, outperforming \cite{christie2020learning} by a wide margin. Comparisons for San Fernando, Argentina (ARG) in Fig. \ref{fig:arg077-height}, Jacksonville, Florida (JAX) in Fig. \ref{fig:jax210-height}, Omaha, Nebraska (OMA) in Fig. \ref{fig:oma285-height}, and Atlanta, Georgia (ATL) in Fig. \ref{fig:atl430-height} all clearly show that our model produces more consistently accurate height predictions and vector fields for rectification, particularly for tall buildings.

$\textbf{Variety of appearance:}$ Our model performs very well for objects that are well-represented in the train set, including tall buildings; however, our model often under-predicts heights for buildings with unique appearance not captured in the train set (\figref{fig:fail-unique-arg-jax}). Failure cases are often viewpoint-dependent, with less accurate predictions for more oblique views and for views without visible shadows (Fig. \ref{fig:fail-viewpoint-arg} and \ref{fig:arg077-height}). To emphasize errors in \figref{fig:fail-small}, we converted RGB images to HSV and replaced intensity with $max(EPE,20)$ normalized to fill the value range, where EPE is endpoint error (pixels).

We believe that more comprehensive geometric augmentations to render novel viewpoints and properly cast shadows may help address this limitation; however, the observed view-dependence of performance suggests that the variety of appearance for building facades must also be addressed. While our initial experiment using multi-view stereo instead of lidar for supervision is limited in scope, we believe that continuing with this approach will help address this challenge of diversity in visual appearance because satellite images can be acquired over much larger scales than lidar.

$\textbf{Partial occlusion:}$ We show anecdotal evidence that smoke from chimneys or smokestacks induce gradual reduction in prediction accuracy (\figref{fig:oma286-smoke-occlusion}). Light haze in images also does not appear to significantly degrade performance (\figref{fig:oma285-height}).

$\textbf{Small vertical features:}$ Our model consistently under-predicts height for small vertical structures (\figref{fig:fail-small}); however, we do not consider this a failure case. We believe that inclusion of these structures in training inhibits learning for larger features that are more relevant to mapping applications, so we remove those reference heights in training. Interestingly, predictions from \cite{christie2020learning} depict the tall antenna shown in \figref{fig:atl430-height}, though predicted heights are inaccurate.

\begin{figure}
    \centering
	\includegraphics[width=0.6\columnwidth]{figs/fail-unique-arg-jax.jpg}
	\caption{Heights of buildings with unusual appearance, highlighted in yellow, are not consistently well predicted.}
	\label{fig:fail-unique-arg-jax}
\end{figure}

\begin{figure}
    \centering
	\includegraphics[width=0.85\columnwidth]{figs/fail-viewpoint-arg.jpg}
	\caption{Prominent failure cases occur for some of the most oblique viewpoints without shadows (shown below).}
	\label{fig:fail-viewpoint-arg}
\end{figure}

\begin{figure}
    \centering
	\includegraphics[width=0.85\columnwidth]{figs/fail-small-vertical-arg-jax.jpg}
	\caption{Small vertical structures such as lamp posts and bridge struts, highlighted yellow, are ignored by our model.}
	\label{fig:fail-small}
\end{figure}

\section*{Acknowledgements}
This work was supported by the National Geospatial-Intelligence Agency and approved for public release, 21-484, with distribution statement A – approved for public release; distribution is unlimited. Commercial satellite images were provided courtesy of DigitalGlobe.

{\small
	\bibliographystyle{ieee_fullname}
	\bibliography{references}
}

\clearpage

\pagebreak
\begin{figure*}
	\includegraphics[width=\textwidth]{figs/arg077-height.jpg}
	\caption{Rectified height images produced using predictions from FLOW-HA \cite{christie2020learning} and our model are compared for ARG tile 31, images 1, 10, and 12. Darker shades of gray represent larger height values. Occluded pixels are blue.}
	\label{fig:arg077-height}
\end{figure*}

\pagebreak
\begin{figure*}
	\includegraphics[width=\textwidth]{figs/jax210-height.jpg}
	\caption{Rectified height images produced using predictions from FLOW-HA \cite{christie2020learning} and our model are compared for JAX tile 210, images 10, 12, and 20. Darker shades of gray represent larger height values. Occluded pixels are blue.}
	\label{fig:jax210-height}
\end{figure*}

\pagebreak
\begin{figure*}
	\includegraphics[width=\textwidth]{figs/oma285-height.jpg}
	\caption{Rectified height images produced using predictions from FLOW-HA \cite{christie2020learning} and our model are compared for OMA tile 285, images 30, 35, and 39. Darker shades of gray represent larger height values. Occluded pixels are blue.}
	\label{fig:oma285-height}
\end{figure*}

\pagebreak
\begin{figure*}
	\includegraphics[width=\textwidth]{figs/atl430-height.jpg}
	\caption{Rectified height images produced using predictions from FLOW-HA \cite{christie2020learning} and our model are compared for ATL tile 430, images 14, 26, and 37. Darker shades of gray represent larger height values. Occluded pixels are blue. Height of the tall antenna is captured in the FLOW-HA height predictions, though without sufficient accuracy for practical rectification. In our training, we apply a median filter to reference lidar height values to ignore tall narrowly occluding features.}
	\label{fig:atl430-height}
\end{figure*}

\begin{figure*}
	\includegraphics[width=\textwidth]{figs/oma286-smoke-occlusion.jpg}
	\caption{Rectified height images produced using predictions from our model are shown for OMA tile 286, images 10-17, with varying seasons and increasing amounts of occluding smoke, top left to bottom right. Heavy smoke results in very inaccurate building height predictions, but performance appears to degrade gradually with the amount of occluding smoke. Darker shades of gray represent larger height values. Occluded pixels are blue.}
	\label{fig:oma286-smoke-occlusion}
\end{figure*}